\def\year{2022}\relax
\documentclass[letterpaper]{article} 
\usepackage{aaai22}  
\usepackage{times}  
\usepackage{helvet}  
\usepackage{courier}  
\usepackage[hyphens]{url}  
\usepackage{graphicx} 
\usepackage{natbib}  
\usepackage{caption} 
\DeclareCaptionStyle{ruled}{labelfont=normalfont,labelsep=colon,strut=off} 
\usepackage{algorithm}
\usepackage{algorithmic}
\usepackage{amsmath}
\usepackage{newfloat}
\usepackage{listings}
\floatstyle{ruled}
\newfloat{listing}{tb}{lst}{}
\floatname{listing}{Listing}
\usepackage{graphicx}

\title{PSG: Prompt-based Sequence Generation for Acronym Extraction}
\author{
	Bin Li\textsuperscript{\rm 1}\equalcontrib,
	Fei Xia\textsuperscript{\rm 2,3}\equalcontrib,
	Yixuan Weng\textsuperscript{\rm 2},
	Xiusheng Huang\textsuperscript{\rm 2,3},
	Bin Sun\textsuperscript{\rm 1},
	Shutao Li\textsuperscript{\rm 1}\thanks{Corresponding Author.}
}
\affiliations{
	\textsuperscript{\rm 1} College of Electrical and Information Engineering, Hunan University\\
	\textsuperscript{\rm 2} National Laboratory of Pattern Recognition, Institute of Automation, Chinese Academy Sciences\\
	\textsuperscript{\rm 3} School of Artificial Intelligence, University of Chinese Academy of Sciences\\
	
	\{libincn, shutao\_li, sunbin611\}@hnu.edu.cn 
	\\
	\{xiafei2020, huangxiusheng2020\}@ia.ac.cn, \text wengsyx@gmail.com
}

\begin{document}

\maketitle

\begin{abstract}
Acronym extraction aims to find acronyms (i.e., short-forms) and their meanings (i.e., long-forms) from the documents, which is important for scientific document understanding (SDU@AAAI-22) tasks. Previous works are devoted to modeling this task as a paragraph-level sequence labeling problem. However, it lacks the effective use of the external knowledge, especially when the datasets are in a low-resource setting. Recently, the prompt-based method with the vast pre-trained language model can significantly enhance the performance of the low-resourced downstream tasks. In this paper, we propose a \textbf{P}rompt-based \textbf{S}equence \textbf{G}eneration (\textbf{PSG}) method for the acronym extraction task. Specifically, we design a template for prompting the extracted acronym texts with auto-regression. A position extraction algorithm is designed for extracting the position of the generated answers. The results on the acronym extraction of Vietnamese and Persian in a low-resource setting show that the proposed method outperforms all other competitive state-of-the-art (SOTA) methods.
\end{abstract}
\section{Introduction}
With the development of technology and global informatization, the number of acronyms is increasing rapidly. The forms of the acronyms are complicated and changeable because of various languages \cite{veyseh2020acronym}. Therefore, understanding the acronyms of long technical phrases is very important for scientific document understanding (SDU@AAAI-22) \cite{veyseh2020does}.
\par
In addition, the carefully designed document reading system should be able to recognize the correct meaning of acronyms and their long forms so that these documents can be processed correctly. This is fairly critical for a variety of downstream tasks, such as question and answer \cite{dijkstra2006question}, reading comprehension \cite{qiu2019survey}, translation \cite{yang2020survey}, medical consultant \cite{li2021more}, etc.
\par
\begin{figure}[t]
	\centering
	\includegraphics[scale=0.36]{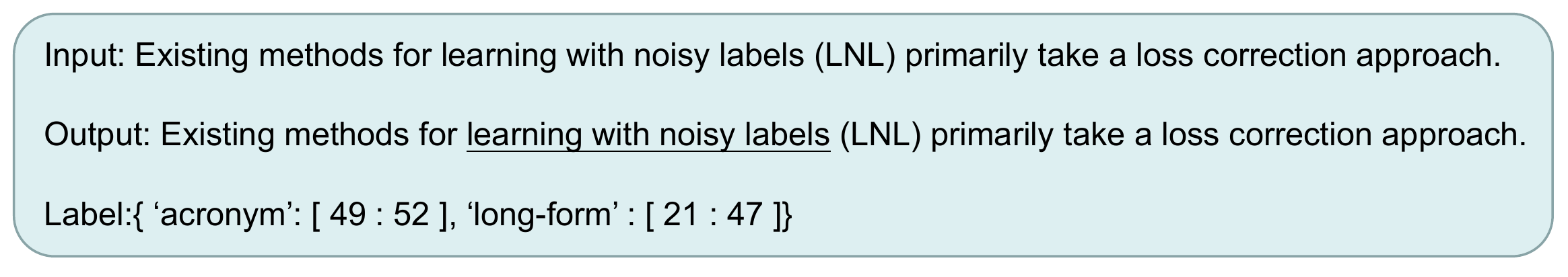}
	\caption{Examples of English acronym extraction.}
	\label{fig1}
	\vspace{-0.4cm}
\end{figure}

The acronym extraction task is mainly used to extract acronyms (i.e., short forms) and their meanings (i.e., long forms) in the science document \cite{veyseh-et-al-2022-Multilingual}. To take the English language as an example, which is shown in Figure \ref{fig1}, the output label is the position of the input text string. In the earliest attempts, the rules or features \cite{schwartz2002simple} are adopted to capture the acronyms. However, these methods often require a lot of manual design, making it hard to process documents with complex grammatical structures.
\par
Recent works tend to model this task as a sequence labeling task \cite{luo2018attention, zhao2020spanmlt, huang-etal-2021-named}, which helps the model to capture the local definition of acronyms in the document. However, the form of acronyms not only appears in the English scene but also in other language scenes (multi-lingual). Traditional sequence labeling methods are weak in utilizing external knowledge \cite{liu2018empower}. As a result, once the system runs on low-resource downstream tasks, its performance is relatively poor. Furthermore, the traditional sequence labeling method requires a lot of manual rules for labeling, which is unrealistic in low-resource scenarios. Inspired by the prompt-based method \cite{liu2021pre}, the prompt tends to extract the knowledge from the large-scale pre-trained model \cite{zheng2021exploring}, which is helpful to improve the generalization ability in low-resource scenarios.
\par 
In this paper, we propose the \textbf{P}rompt-base \textbf{S}equence \textbf{G}eneration (\textbf{PSG}) method for acronym extraction. Specifically, we design a prompt for sequence generation with the large-scale pre-trained model. With the prompt being used as a template, the acronym form and the long-form can be generated via auto-regression. After obtaining the answer, we designed a position extraction algorithm to locate the labels. The proposed method ranks 1-st under the low-resource language setting (i.e., Vietnamese and Persian) in the shared task 1 of the SDU@AAAI-22, which outperforms all other competitive methods.
The main contributions are summarized as follows:
\begin{itemize}
\item As far as we know, this is the first attempt to adopt prompt-based sequence generation for the acronym extraction task.
\item We propose a novel acronym extraction method, including prompt-based sequence generation method and position extraction algorithm for obtaining the final labels.
\item Extensive experiments are conducted on the low-resource datasets (i.e., Vietnamese and Persian). The results demonstrate the effectiveness of our proposed method compared with other competitive baselines.
\end{itemize}
\section{Task introduction}
\subsection{Problem definition}
We treat the acronym task as a sequence generation problem. Given a series of tokens in the text $\textbf{x} = \{x_{1}, x_{2}, \ldots, x_{n}\}$, this task aims at finding the corresponding position from the original text. The label indicates the short-form $S$ (i.e., acronym) and the long-form $L$ (i.e., phrase). We formulate the above process as follows:
\begin{equation}
L, S=h\left(x_{1}, x_{2}, \ldots, x_{n}\right)
\end{equation}
where $h$ is the model which extracts the answers.
\subsection{Evaluation metric}
The submitted results will be evaluated based on the macro-averaged precision, recall, and F1 scores on the online test set. The final scores represent the prediction correctness of short-form (i.e., acronym) and long-form (i.e., phrase) boundaries in the sentences. 
The short-form or long-form boundary prediction is counted as correct if the beginning and the end of the predicted short-form or long-form boundaries are equal to the ground-truth beginning and end of the short-form or long-form boundary, respectively. 
The official score is the macro average of short-form and long-form F1 scores.
\subsection{Dataset}
\begin{table}[h]
	\centering
	\renewcommand\arraystretch{1.2}
	\caption{Statistical Information of Vietnamese Dataset.}
	\begin{tabular}{c|cc}
			\noalign{\hrule height 1pt}
		\textbf{Data} & \textbf{Sample Number} & \textbf{Ratio} \\ 
			\noalign{\hrule height 0.5pt}
		\textbf{Training Set}       & 1274        & 79.98\%         \\ 
		\textbf{Development Set}      & 159        & 9.98\%         \\ 
		\textbf{Test Set}       & 160       & 10.04\%         \\ 			\noalign{\hrule height 0.5pt}
		\textbf{Total}      & 1593         & 100\%         \\ 			\noalign{\hrule height 1pt}
	\end{tabular}
	\label{table1}
\end{table}
\vspace{-0.3cm}
\begin{table}[h]
		\renewcommand\arraystretch{1.2}
	\centering
		\caption{Statistical Information of Persian Dataset.}
	\begin{tabular}{c|cc}
		\noalign{\hrule height 1pt}
		\textbf{Data} & \textbf{Sample Number} & \textbf{Ratio} \\ 
		\noalign{\hrule height 0.5pt}
		\textbf{Training Set}       & 1336        & 80.34\%         \\ 
		\textbf{Development Set}      & 167        & 10.04\%         \\ 
		\textbf{Test Set}       & 160       & 9.62\%         \\ \noalign{\hrule height 0.5pt}
		\textbf{Total}      & 1663         & 100\%         \\ \noalign{\hrule height 1pt}
	\end{tabular}
	\label{table2}
	\vspace{-0.1cm}
\end{table} 
This acronym extraction task consists of various multi-lingual datasets composed of document sentences in science fields \cite{veyseh-et-al-2022-MACRONYM}. Among them, the Vietnamese dataset and the Persian dataset are set in a low-resource scenario compared to other languages. 
As shown in Table \ref{table1}, the Vietnamese dataset is divided into training (1274), development (159), and testing according to the data set (160).
As shown in Table \ref{table2}, the Persian dataset is divided into training (1336), development (167), and testing according to the data set (160). 
The training and validation sets of the above two datasets have been manually labeled, where the label is a list of position boundaries. 
\section{Method}
In this section, we will introduce our method in detail, including the model architecture, prompt design, sequence generation and position extraction algorithm.
\begin{figure}[h]
	\centering
	\includegraphics[scale=0.8]{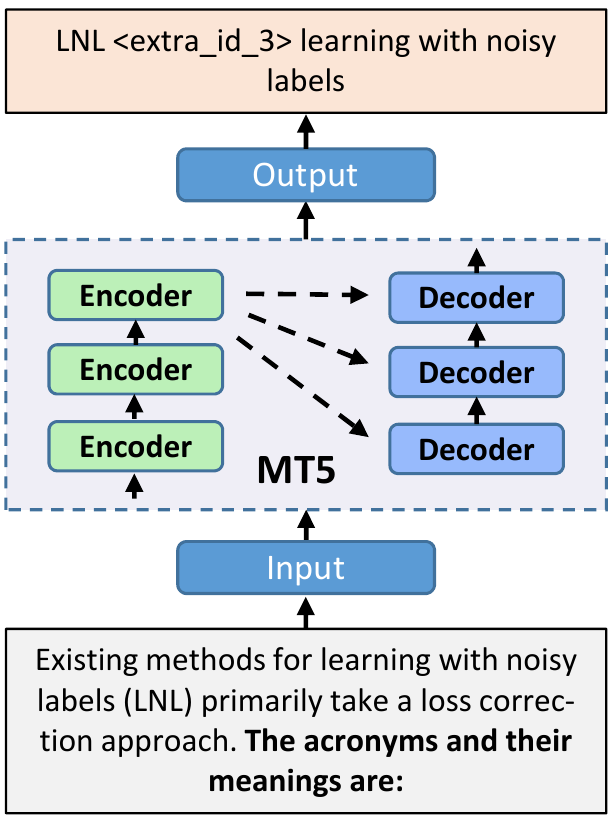}
	\caption{Overview of model architecture, where the sentence in bold is the prompt and the $<$extra\_id\_3$>$ is designed to separate the acronym of short-form and long-form.} 
	\label{fig2}
	\vspace{-0.2cm}
\end{figure}
\subsubsection{Model architecture}
The overall architecture of our me-thod is shown in Figure \ref{fig2}, the MT5 model \cite{xue2021mt5} is adopted as our backbone for sequence generation. We first input the text with a manually designed prompt to be tokenized with MT5 tokenizer, then the input shall be encoded with the encoder through the self-attention \cite{vaswani2017attention} mechanism. Finally, the output is produced by the decoder via auto-regression. Notice that the output contains the unused token, which is designed as the placeholder for prompt tuning, thus further utilizing the external knowledge from the pre-trained model. 
\subsubsection{Prompt design}
We manually design the prompt to extract relevant knowledge from the pre-trained model for sequence generation, which is presented as the fixed tokens, i.e., ``The acronyms and their meanings are:''. In addition, the unused tokens are adopted as a placeholder to control the outputs. Specifically, the unused tokens are used as the placeholder to form a template for prompt tuning, where the $<$extra\_id\_1$>$ represents the separator of the short forms, the $<$extra\_id\_2$>$ represents the separator of the long forms, and the $<$extra\_id\_3$>$ represents the separator between the acronym of the long-form and short-form. The $<$extra\_id\_4$>$ indicates that no acronym of short-form appears, while the $<$extra\_id\_5$>$ indicates that no acronym of long-form appears.
\subsubsection{Sequence generation}
The sequence generation task is designed to generate extracted possible acronyms. The text $\textbf{x} = (x_{1}, x_{2}, \ldots, x_{n})$ is encoded through the encoder to obtain the context encoding $E_{C}$. At the decoding stage, the loss function of the sequence generation task can be performed as auto-regression, which is shown as equation (\ref{sequence_generation}):
\begin{equation}
L_{D}(\varphi) =-\sum_{i} \log P_{\varphi}\left({y}_{i} \mid {y}_{0}, \ldots, {y}_{i-1}, E_{C}\right) 
\label{sequence_generation}
\end{equation}
where the $\varphi$ is the parameters of the model, the $y_i$ represents the \textit{i}-th word generated by the decoder, and ${y_{0}, \ldots,y_{i-1}}$ is a sequence of previously generated tokens.
\subsubsection{Position extraction algorithm}
After the possible acron-yms of short-form and long-form are generated with sequence generation, the next step is to consider how to extract their positions. A good extraction algorithm determines the final quality of the generated result. We use a greedy traversal search method,  adopting the regular method from left to right to find the corresponding location boundary. At the same time, we need to ensure there is no overlap in the extracted outputs by detecting the boundary margins so that the extracted positions are independent of each other. To take the acronym of short-form as an example, the algorithm is represented in Algorithm \ref{algo}.
\begin{algorithm}
	\caption{Position extraction algorithm}
	\label{alg:algorithm}
	\textbf{Input}: Output text $\textbf{y}$ tokenized into list $L$; Original input text $\textbf{x}$\\
	\textbf{Output}: \leftline{ Short-form acronym position list $D$}
	\vspace{-0.4cm}
	\begin{algorithmic}[1] 
		\STATE Let $D$ = []
		\STATE Let truncated text $\textbf R$ = $\emptyset$
		\IF {`$<$extra\_id\_4$>$' in $L$}
		\STATE $D$ = []
		\ELSE
		\STATE $i$ =  $\textbf{y}$.find(`$<$extra\_id\_3$>$')
		\STATE $\textbf R$ = $\textbf{y}$[:$i$].strip() 
    	\FOR{$j$ in $\textbf R$.split(` $<$extra\_id\_1$>$ ')} 
    		\FOR{$m$ in [$k$.start() for $k$ in re.finditer($j$, $\textbf{x}$)]} 
				\STATE $D$.append([$m$, $m$ + len($j$)])  	
			\ENDFOR
		\ENDFOR
		\ENDIF
		\STATE Making the position intervals in list $D$ independent of each other
		\STATE \textbf{return} solution $D$
	\end{algorithmic}
	\label{algo}
\end{algorithm}
	\vspace{-0.3cm}
\section{Experiment setup}
\subsection{Baseline models}
\begin{itemize}
\item \textbf{Rule-based method} The baseline method proposed by Schwartz is a rule-based
method \cite{schwartz2002simple}. In this baseline, the words that more than 60\% of their characters are uppercased are selected as acronym. To select long-forms, if the initial characters of the preceeding words before an acronym can form the acronym they are selected as long-form. The related codes can be found on the website\footnote{https://github.com/amirveyseh/AAAI-22-SDU-shared-task-1-AE}.
\item \textbf{BERT-CRF model}
The BERT-CRF \cite{luo2018attention} architecture is composed of a BERT model \cite{bert} concatenated with a token-level perception layer with a conditional random field (CRF) on top. For the input tokens, the BERT model produced encoded tokens and the classification model projects encodings to the label space. The classification output scores are then sent to the CRF layer, whose parameters are the tag transition matrix, where the elements represent the tag transition score. The matrix contains two states: begin (B) and end (E). We only consider the cross-entropy loss of the first sub-token of each token.
\item \textbf{BERT-Span model}
The task is also considered as the boundary of phrase spans modeled by BERT-Span model \cite{zhao2020spanmlt}, including acronyms of short-form and long-form. Two binary classifiers are adopted to output the multiple start and end indexes. The prediction represents whether each token is the start or end tag. Given each token representation from the BERT model, the probabilities of each token are predicted as the start or end position. The weighted cross-entropy loss is implemented to train the model with parameters shared at the BERT encoder layers.
\item \textbf{MT5 model} The MT5 \cite{xue2021mt5} is a multilingual Transformer model pre-trained on a dataset (mC4) containing text from 101 different languages include the language of Vietnamese and Persian. The architecture of the MT5 model, which is based on T5 \cite{raffel2019exploring}, is designed to support any Natural Language Processing task by reframing the required task as a sequence-to-sequence task. Also, the MT5 model has different variants to perform the sequence generation task, where we final adopt the model size of the base, the large and the x-large. The above models can be found and downloaded on the website\footnote{https://huggingface.co/models}.
\end{itemize}
\begin{table*}
	\hspace{1.2em}
	\begin{minipage}{\textwidth}
		\begin{minipage}[h]{0.4\textwidth}
			\centering
			\makeatletter\def\@captype{table}
			\makeatother\caption{F1 Performace in Vietnamese.}			
			\renewcommand\arraystretch{1.2}	\setlength{\tabcolsep}{4mm}
			\begin{tabular}{ccc} 
				\noalign{\hrule height 1pt}
				Method              & Val F1 & Test F1 \\ \noalign{\hrule height 0.5pt}
				Rule-based          & 0.5337       & 0.5646             \\ 
				BERT-CRF            & 0.7712       & 0.7613        \\ 
				BERT-Span           & 0.8118       & 0.7844        \\ 
				MT5-base            & 0.8222       & 0.8012            \\ 
				\noalign{\hrule height 0.5pt}
				PSG-base          & 0.8313       & 0.8195        \\  
				PSG-large         & 0.8523       & 0.8344        \\ 
				PSG-xlarge & \textbf{0.8611}       & \textbf{0.8416}        \\ \noalign{\hrule height 1pt}
			\end{tabular}
			\label{Vie}
		\end{minipage}
		\begin{minipage}[h]{0.66\textwidth}
			\centering
			\makeatletter\def\@captype{table}
			\makeatother\caption{F1 Performace in Persian.}		
			\renewcommand\arraystretch{1.2}	 		\setlength{\tabcolsep}{4mm}
			\begin{tabular}{ccc} 
				\noalign{\hrule height 1pt}
				Method              & Val F1 & Test F1 \\ \noalign{\hrule height 0.5pt}
				Rule-based          & 0.5811       & 0.5636             \\ 
				BERT-CRF            & 0.6126       & 0.5782        \\ 
				BERT-Span           & 0.6334       & 0.5916        \\ 
				MT5-base            & 0.6578       & 0.6613             \\ 
				\noalign{\hrule height 0.5pt}
				PSG-base          & 0.6901      & 0.7077         \\  
				PSG-large         & 0.7363       & 0.7437        \\ 
				PSG-xlarge & \textbf{0.7745}       & \textbf{0.7993}       \\ \noalign{\hrule height 1pt}
			\end{tabular}
			\label{Per}
		\end{minipage}
	\end{minipage}
\end{table*}
\subsection{Training strategies}
For the sequence generation part, we design a curriculum learning method to train the sequence generation model, where the model is fine-tuned with a corpus of different difficulties. We mix up all the training datasets in the acronym tasks, including 4,000 English, 1,000 Persian, and 800 Vietnamese paragraphs in the scientific domain and 4,000 English, 8,000 French, 6,400 Spanish, and 3,000 Danish paragraphs in the legal domain. The training  steps are as follows:
\begin{enumerate}
	\item The trained MT5 model is utilized to initialize the parameters of the encoder and decoder, and fine-tune with the multi-lingual data. Finally, we train the model with 8 epochs.
	\item We fine-tune the MT5 model with a single language for Vietnamese and Persian respectively with 8 epochs being trained.
\end{enumerate}
\par Notice that the Persian is a right-to-left language, it is different from other languages. However, we deem that the whole training process is an auto-regressive task, where the pre-trained model can learn the features of multi-languages via self-supervised pre-training and fine-tuning. In short, the model can well produce the results with the proposed position extraction algorithm.
\subsection{Implementation}
All models are implemented based on the open-source transformers library of huggingface \cite{wolf2019huggingface}, where thousands of pretrained models are provided to perform different tasks on texts such as sequence labeling and sequence generation. The huggingface toolkit provides multiple APIs to quickly download and use those pre-trained models, thus fine-tuning them on the downstream tasks. We adopt Pytorch deep learning framework to finish this task. Specifically, we use four GPUs of NVIDIA 3090 with 24 cores to complete these experiments. 
\par
In BERT-CRF method, we initialize the model with mbert \cite{libovicky2019language}, and initial learning rates are 5e-5 and 5e-2 for BERT and CRF respectively. We utilize the AdamW optimizer \cite{loshchilov2017decoupled} with a batch size of 32. 
\par
In BERT-Span method, we initialize the model with mbert, and initial learning rates are 5e-5. We utilize the AdamW optimizer with a batch size of 32. 
\par
Our models are implemented with various varients, where the base model is implemented with the batch size of 32, the large model is implemented with the batch size of 4 and the xlarge model is implemented with the batch size of 2. 
\par
As for multi-lingual fine-tune and single language fine-tuning,  we used the AdamW optimizer with an initial learning rate of 1e-4 and annealed it gradually after a warm-up epoch until it reached 1e-5.
\section{Results}
The main results of our model and baselines are shown
in Table \ref{Vie} and Table \ref{Per}, where the F1 performance in Vietnamese and Persian are presented respectively. It can be found that the pre-trained model has more advantages than the rule-based method, since the rule-based method has limited generalization capabilities in validation and test datasets. The BERT-CRF and BERT-Span methods have similar performance. This may be because the sequence labeling method with pre-trained model has limited ability to capture external information in the unseen test dataset, especially in the low-resource setting. It is worth noting that the F1 scores of the method based on the generation method are higher than the sequence labeling method, which indicates that the generation method is more conducive to capturing the relationship between the acronym of the short-form and the long-form. What's more, our method is higher than other baselines in both the validation set and the test set. More precisely, on the Vietnamese test set, the proposed method reaches an F1 score of 0.8416, and on the Persian test set, the proposed method reaches a score of 0.7993. Further conclusion can be found that: 1) compared with the traditional generation method (i.e., MT5), we have designed the prompt method, which is effective to make full use of the knowledge from the pre-trained model. As a result, the proposed method gets a good performance on the test set as its stronger generalization ability. 2) As the scale of the initial pre-trained model increases, the prompt can enhance the generalization of downstream tasks.
\section{Conclusion}
We describe the sequence generation system used for submission in the acronym extraction task of SDU@AAAI-22. Three types of methods have been tried to complete this task, including the rule-base, sequence labeling, and sequence generation. The results show that the proposed method outperforms the other baselines, achieving the best performance (Top-1) in the acronym extraction of Vietnamese and Persian. It can be concluded that the sequence generation method has more ability for generalization than the sequence labeling. The prompt can further enhance the generalization ability. Furthermore, there are some promising works to be done in the future. As for the training method, it will be a wise way to merge the sequence labeling into the sequence generation as the multi-task learning. As for the prompt design, the soft prompt can be a better way to bridge the gap between the upstream and the downstream task than the fixed prompt templates.
\section*{Acknowledgement}
{This work is supported by the National Key Research and Development Project of China (2018YFB1305200) and the National Natural Science Fund of China (62171183, 61801178).}

\bibliography{references}
\end{document}